\pdfoutput=1
\documentclass[sigconf]{acmart}
\AtBeginDocument{%
  \providecommand\BibTeX{{%
    \normalfont B\kern-0.5em{\scshape i\kern-0.25em b}\kern-0.8em\TeX}}}

\copyrightyear{2020}
\acmYear{2020}
\setcopyright{acmcopyright}
\acmConference[MM '20]{Proceedings of the 28th ACM International Conference on Multimedia}{October 12--16, 2020}{Seattle, WA, USA}
\acmBooktitle{Proceedings of the 28th ACM International Conference on Multimedia (MM '20), October 12--16, 2020, Seattle, WA, USA}
\acmPrice{15.00}
\acmDOI{10.1145/3394171.3413827}
\acmISBN{978-1-4503-7988-5/20/10}

\usepackage{epsfig}
\usepackage{graphicx}
\usepackage{amsmath}
\usepackage{amssymb}
\usepackage{subfigure}
\usepackage{multirow}
\usepackage{tabularx}
\usepackage{xcolor}
\usepackage{bm}
\usepackage{soul}

\DeclareMathOperator*{\argmin}{arg\,min}

\begin{document}
\fancyhead{}
\title{Uncertainty-based Traffic Accident Anticipation with Spatio-Temporal Relational Learning}

\author{Wentao Bao}
\email{wb6219@rit.edu}
\orcid{0000-0003-2571-3341}
\affiliation{%
  \institution{Rochester Institute of Technology}
  \city{Rochester}
  \state{New York}
  \country{USA}
}

\author{Qi Yu}
\email{qi.yu@rit.edu}
\affiliation{%
  \institution{Rochester Institute of Technology}
  \city{Rochester}
  \state{New York}
  \country{USA}
}

\author{Yu Kong}
\email{yu.kong@rit.edu}
\affiliation{%
  \institution{Rochester Institute of Technology}
  \city{Rochester}
  \state{New York}
  \country{USA}
}

\def\etal{et~al.~}

\begin{abstract}
  Traffic accident anticipation aims to predict accidents from dashcam videos as early as possible, which is critical to safety-guaranteed self-driving systems. With cluttered traffic scenes and limited visual cues, it is of great challenge to predict how long there will be an accident from early observed frames. Most existing approaches are developed to learn features of accident-relevant agents for accident anticipation, while ignoring the features of their spatial and temporal relations. Besides, current deterministic deep neural networks could be overconfident in false predictions, leading to high risk of traffic accidents caused by self-driving systems. In this paper, we propose an uncertainty-based accident anticipation model with spatio-temporal relational learning. It sequentially predicts the probability of traffic accident occurrence with dashcam videos. Specifically, we propose to take advantage of graph convolution and recurrent networks for relational feature learning, and leverage Bayesian neural networks to address the intrinsic variability of latent relational representations. The derived uncertainty-based ranking loss is found to significantly boost model performance by improving the quality of relational features. In addition, we collect a new Car Crash Dataset (CCD) for traffic accident anticipation which contains environmental attributes and accident reasons annotations. Experimental results on both public and the newly-compiled datasets show state-of-the-art performance of our model. Our code and CCD dataset are available at: \href{https://github.com/Cogito2012/UString}{https://github.com/Cogito2012/UString}.
\end{abstract}



\begin{teaserfigure}
  \includegraphics[width=\textwidth]{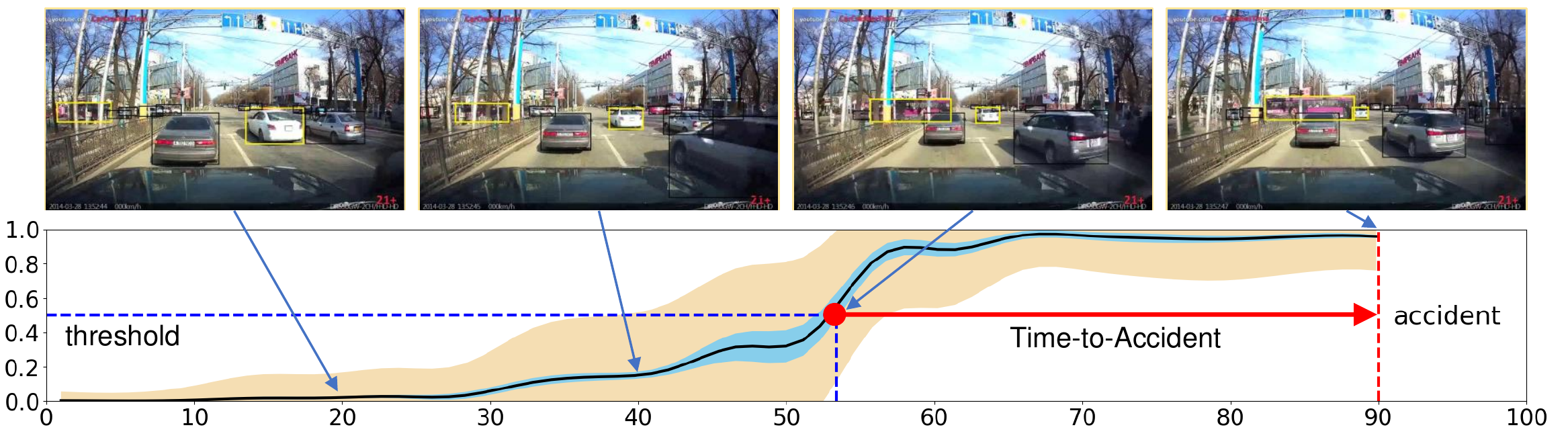}
  \caption{Illustration of Uncertainty-based Accident Anticipation. This paper presents a novel model to predict the probabilities (black curve) of a future accident (ranges from 90-th to 100-th frame). Our goal is to achieve early anticipation (large Time-to-Accident) giving a threshold probability (horizontal dashed line), while estimating two kinds of predictive uncertainties, i,e., aleatoric uncertainty (wheat color region) and epistemic uncertainty (blue region).}
  \label{fig:teaser}
\end{teaserfigure}

\maketitle

\section{Introduction}
Accident anticipation aims to predict an accident from dashcam video before it happens. It is one of the most important tasks for safety-guaranteed autonomous driving applications and has been receiving increasing attentions in recent years~\cite{ChanACCV2016,SuzukiCVPR2018,FangITSC2019,CorcoranCRV2019}. Thanks to accident anticipation, the safety level of intelligent systems on vehicles could be significantly enhanced. For example, even a successful anticipation made with only a few seconds earlier before the accident happens can help a self-driving system to make urgent safety control, avoiding a possible car crash accident.

However, accident anticipation is still an extremely challenging task due to noisy and limited visual cues in an observed dashcam video. Take Fig.~\ref{fig:teaser} as an example, a traffic scene captured in egocentric view is typically crowded with multiple cars, pedestrians, motorcyclists, and so on. In this scenario, accident-relevant visual cues could be overwhelmed by objects that are not relevant to the accident, making an intelligent system insensible to a car crash accident happened at the road intersection. Nevertheless, traffic accidents are foreseeable by training a powerful uncertainty-based model to distinguish the accident-relevant cues from noisy video data. For example, the inconsistent motions of multiple vehicles may indicate high risk of possible future accidents.

In this paper, we propose a novel uncertainty-based accident anticipation model with spatio-temporal relational learning. The model aims to learn accident-relevant cues for accident anticipation by considering both spatial and temporal relations among candidate agents. The candidate agents are a group of moving objects like vehicles and their relational features are indicative of future unobserved accidents. The spatial relations of candidate agents are learned from their spatial distance, visual appearance features, as well as historical visual memory. The temporal relations of agents provide learnable patterns to indicate how the agents evolve and end with an accident in temporal context. It can be recurrently learned by updating historical memory with agent-specific features and the spatial relational representation. To address the variability of the spatio-temporal relational representations, a probabilistic module is incorporated to simultaneously predict accident scores and estimate how much uncertainty when making the prediction.

As shown in Fig.~\ref{fig:framework}, on one hand, we propose to learn spatial relations with graph convolutional networks (GCN)~\cite{DefferrardNIPS2016,KipfICLR2017} by considering the hidden states from recurrent neural network (RNN))~\cite{LiptonArXiv2015,SeoICLR2017} cell. On the other hand, we propose to build temporal relations with RNNs by considering both spatial relational and agent-specific features. The cyclic process of the coupled GCNs and RNNs could generate representative latent spatio-temporal relational features. Besides, we propose to incorporate Bayesian deep neural networks (BNNs)~\cite{DenkerNIPS1990,NealBook2012} into our model to address the predictive uncertainty. With the Bayesian formulation, our derived epistemic uncertainty-based ranking loss is effective to improve the quality of the learned relational features and significantly leads to performance gain. At last, to further consider the global guidance of all hidden states in training stage, we propose a self-attention aggregation layer as shown in Fig.~\ref{fig:saa}, from which an auxiliary video-level loss is obtained and demonstrated beneficial to our model.

Compared with existing RNN-based methods~\cite{ChanACCV2016,SuzukiCVPR2018}, our model captures not only agent-specific features but also relational features for accident anticipation. Compared with the recent approach~\cite{NeumannCVPRW2019} which is developed with 3D CNNs, our model is developed with GCNs and RNNs so that both spatial and temporal relations can be learned. Moreover, our method is capable of estimating the predictive uncertainty while all existing methods are deterministic.

The proposed model is evaluated on two public dashcam video datasets, i.e., DAD~\cite{ChanACCV2016} and A3D~\cite{SuzukiCVPR2018}, and our collected Car Crash Dataset (CCD). Experimental results show that our model can outperform existing methods on all datasets. For DAD datasets, our method can anticipate traffic accident 3.53 seconds on average earlier before an accident happens. With best precision setting, our model can achieve 72.22\% average precision. Compared with DAD and A3D datasets, our CCD dataset includes diversified environmental annotations and accident reason descriptions, which could promote research on traffic accident reasoning.

The main contributions of this paper are summarized below:
\begin{itemize}
    \item We propose a traffic accident anticipation model by considering both agent-specific features and their spatio-temporal relations, as well as the predictive uncertainty.
    \item With Bayesian formulation, the spatio-temporal relational representations can be learned with high quality by a novel uncertainty-based ranking loss.
    \item We propose a self-attention aggregation layer to generate video-level prediction in the training stage, which serves as global guidance and is demonstrated beneficial to our model.
    \item We release a new dataset containing real traffic accidents, in which diversified environmental annotations and accident reasons are provided.
\end{itemize}

\section{Related Work}

\subsection{Traffic Accident Anticipation}

To anticipate traffic accidents that happened in future frames, an intuitive solution is to iteratively predict accident confidence score for each time step. Chan~\etal~\cite{ChanACCV2016} recently proposed DSA framework to leverage candidate objects appeared in each frame to represent the traffic status. They applied spatial-attention on these objects to get weighted feature representation for each LSTM cell. Based on this work, Suzuki~\etal~\cite{SuzukiCVPR2018} proposed an adaptive loss for early anticipation with quasi-recurrent neural networks~\cite{BradburyICLR2017}. Similar to DSA that implements dynamic-spatial attention to focus on accident-relevant objects, Corcoran and James~\cite{CorcoranCRV2019} proposed a two-stream approach to traffic risk assessment. They utilized features of candidate objects as spatial stream and optical flow as temporal stream, and the two-stream features are fused for risk level classification. Instead of using dashcam videos, Shah~\etal~\cite{ShahT4SW2018} proposed to use surveillance videos to anticipate traffic accidents by using the framework DSA. Different from previous works, recently Neumann and Zisserman~\cite{NeumannCVPRW2019} used 3D convolutional networks to predict the sufficient statistics of a mixture of 1D Gaussian distributions. In addition to using only dashcam video data, Takimoto~\etal~\cite{TakimotoPredictGIS2019} proposed to incorporate physical location data to predict the occurrence of traffic accidents. Closely related to traffic accident anticipation, the traffic accident detection is recently studied by Yao~\etal~\cite{YaoIROS2019}. They proposed to detect traffic anomalies by predicting the future locations on video frames using ego-motion information. To anticipate both spatial risky regions and temporal accident occurrence, Zeng~\etal~\cite{ZengCVPR2017} proposed a soft-attention RNN by considering event agent such as human that triggers the event.

However, existing work typically ignores the relations between accident-relevant agents which capture important cues to anticipate accidents in future frames. Besides, none of them considers the uncertainty estimation in developing their models, which is critical to safety-guaranteed systems.

\subsection{Uncertainty in Sequential Modeling}

Uncertainty estimation is crucial to sequential relational modeling. One way is to directly formulate the latent representations of relational observations at each time step as random variables, which follow posterior distributions that can be approximated by deep neural networks. This is similar to variational auto-encoder (VAE)~\cite{KingmaICLR2014,RezendeICML2014}. Inspired by VAE, Chung~\etal~\cite{ChungNIPS2015} proposed variational recurrent neural network (VRNN) which formulates the hidden states of RNN as random variables and uses neural networks to approximate the posterior distributions of the variables. To further consider the relational representation of sequential data, Hajiramezanali~\etal~\cite{HajiramezanaliNIPS2019} proposed variational graph recurrent neural networks (VGRNN) for dynamic link prediction problem by combining the graph RNN and variational graph auto-encoder (VGAE)~\cite{KipfNIPSW2016}.

Another way to address uncertainty estimation is to formulate the weights of neural network as random variables such as Bayesian neural networks (BNNs)~\cite{DenkerNIPS1990,NealBook2012}. Recently, Zhao~\etal~\cite{ZhaoICCV2019} proposed a Bayesian graph convolution LSTM model for skeleton-based action recognition. In this paper, we also use graph convolution and BNNs but the difference is that their method uses stochastic gradient Hamiltonian Monte Carlo (SGHMC) sampling for posterior approximation, while we use Bayes-by-backprop~\cite{BlundellJMLR2015} as our approximation method. Compared with SGHMC, Bayes-by-backprop can be seamlessly integrated into deep learning optimization process so that it is more flexible to handle the learning tasks with large-scale dataset, i.e., dashcam videos used in traffic accident anticipation.

\begin{figure}
    \centering
    \includegraphics[width=\linewidth]{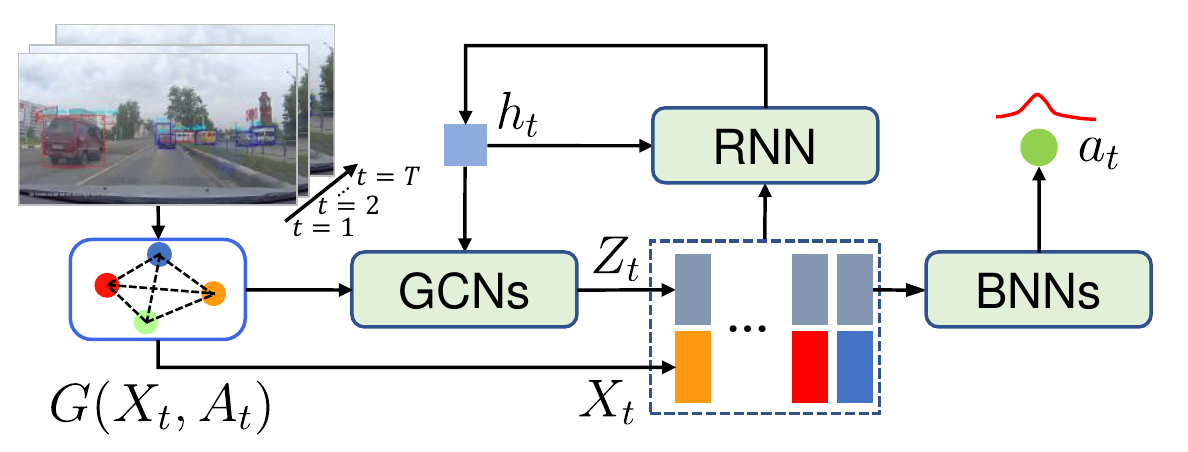}
    \caption{Framework of the proposed model. With graph embedded representations $G(\bm{X}_t, \bm{A}_t)$ at time step $t$, our model learns the latent relational representations $\bm{Z}_t$ by the cyclic process of graph convolutional networks (GCNs) and recurrent neural network (RNN) cell, and predicts the accident score $\bm{a}_t$ by Bayesian neural networks (BNNs).}
    \label{fig:framework}
\end{figure}

\section{Proposed Method}

\textbf{Problem Setup.} In this paper,  the goal of accident anticipation is to predict an accident from dashcam videos before it happens. Formally, given a video with current time step $t$, the model is expected to predict the probability $a_{t}$ that an accident event will happen in the future.
Furthermore, suppose an accident will happen at time step $y$ where $t<y$, the \emph{Time-to-Accident} (TTA) is defined as $\tau = y - t$ when $t$ is the first time that $a_{t}$ is larger than given threshold (see Fig.~\ref{fig:teaser}). For any $t\ge y$ with a positive video that contains an accident, we define $\tau=0$ which means the model fails to anticipate the accident. In this paper, our goal is to predict $a_{t}$ and expect $\tau$ to be as large as possible for dashcam videos that contain accidents. Similar to~\cite{ChanACCV2016}, the ground truth of $a_{t}$ is expressed with 2-dimensional one-hot encoding so that prediction target is $\bm{a}_t=(a_t^{(p)},a_t^{(n)})^T$, where $a_t^{(p)}$ and $a_t^{(n)}$ represent the positive and negative predictions, respectively, meaning an accident will happen or not happen in the given video.

\textbf{Framework Overview.} The framework of our model is depicted in Fig.~\ref{fig:framework}. With a dashcam video as input, a graph is constructed with detected objects and corresponding features at each time step. To learn the spatio-temporal relations of these objects, we use graph convolutional networks (GCNs) to learn the spatial relations and leverage the hidden state $\bm{h}_t$ of recurrent neural network (RNN) cell to enhance the input of the last GCN layer. Besides, the latent relational features are fused with corresponding object features as input of an RNN cell to update the hidden state at next time step. The cyclic process encourages our model to learn the latent relational features $\bm{Z}_t$ from both spatial and temporal aspects. Furthermore, we propose to use Bayesian neural network (BNN) to predict accident scores $a^{t}$ so that predictive uncertainties are naturally formulated. During the training stage, we propose a self-attention aggregation (SAA, in Fig.~\ref{fig:saa}) layer to predict video-level score, which can globally guide the learning of the proposed model.

In the following sections, each part of our model will be introduced in detail.

\subsection{Spatio-Temporal Relational Learning}

The spatio-temporal relations of traffic accident-relevant agents are informative to predict future accidents. In our model, we propose to use graph structured data to represent the observation at each time step. Then, the feature learning of spatial and temporal relations are coupled into a cyclic process.

\textbf{Graph Representation.} Graph representation for traffic scene has the advantages over full-frame feature embedding in that the impact of cluttered traffic background can be reduced and informative relations of traffic agents can be discovered for accident anticipation. Similar to~\cite{ChanACCV2016,ShahT4SW2018}, we exploit object detectors~\cite{RenNIPS2015,CaiCVPR2018} to obtain a fixed number of candidate objects. These objects are treated as graph nodes so that a complete graph can be formed. However, the computational cost of graph convolution could be tremendous if the node features are with high dimensions.

In this paper, to construct low-dimensional but representative features for graph nodes $\bm{X}_t$, we introduce fully-connected (FC) layers to embed both the features of full-frame and candidate objects into the same low-dimensional space. Then, the frame-level and all object-level features are concatenated to enhance the feature representation capability:
\begin{equation}
    \bm{X}_t^{(i)} = \left[\Phi \left(\bm{O}_t^{(i)}\right), \Phi \left(\bm{F}_t\right)\right],
\end{equation}
where $\Phi$ denotes FC layer, $\bm{O}_t^{(i)}$ and $\bm{F}_t$ are high-dimensional features of the $i$-th object and corresponding frame at time $t$, respectively. The operator $[,]$ represents concatenation in feature dimension and is used throughout this paper for simplicity.

The graph edge at time $t$ is expressed as an adjacent matrix $\bm{A}_t$ of a complete graph since we do not have information on which candidate object will be involved in an accident. Typically, an object with closer distance to others has higher possibility to be involved in an future accident. Therefore, the spatial distance between objects should be considered in edge weights such that we define $\bm{A}_t$ as
\begin{equation}
    \bm{A}_t^{(ij)}= \frac{\exp\{-d(r_i, r_j)\}}{\sum_{ij} \exp\{-d(r_i, r_j)\}},
\label{eq:edge}
\end{equation}
where $d(r_i,r_j)$ measures the Euclidean distance between two candidate object regions $r_i$ and $r_j$. By this formulation, closer distance leads to larger $\bm{A}_t^{(ij)}$. This means the two objects $i$ and $j$ will be applied with larger weight when we use graph convolution to learn their relational features for accident anticipation. Note that due to object occlusions, small distance defined in pixel space does not necessarily indicate close distance in physical world. It is possible to use 3D real-world distance if camera intrinsics are known. Nevertheless, the adjacency matrix defined in Eq.~\ref{eq:edge} has advantage to suppress the impact of irrelevant objects with significant large pixel distance to relevant objects.

\textbf{Temporal Relational Learning.} To build temporal relations at different time steps, RNN methods such as LSTM~\cite{LSTM} and GRU~\cite{GRU} are widely adopted in existing works. However, traffic objects may not always be remained in each frame, the node features of the statically structured graph will be dynamically changing over time. Thanks to the recent graph convolutional recurrent network (GCRN)~\cite{SeoICLR2017}, it can handle the node dynamics defined over a static graph structure~\cite{HajiramezanaliNIPS2019}. Therefore, we propose to adapt GCRN for temporal relational feature learning. Specifically, the hidden states $\bm{h}_{t}$ of RNN cell at each time step are recurrently updated by
\begin{equation}
    \bm{h}_{t+1} = \text{GCRN}\left(\left[\bm{Z}_t,\bm{X}_t\right], \bm{h}_{t}\right),
    \label{eqn:rnn}
\end{equation}
where $\bm{Z}_t$ is the relational feature generated by the last GCN layer. The feature fusion between $\bm{Z}_t$ and $\bm{X}_t$ ensures our model to make fully use of both agent-specific and relational features.

\textbf{Spatial Relational Learning.} To capture spatial relations of detected objects, we follow the graph convolution defined by~\cite{DefferrardNIPS2016,KipfICLR2017} for each GCN layer. In this paper, we use two stacked GCN layers and consider the hidden state $\bm{h}_t$ learned by RNNs to learn the spatial relational features:
\begin{equation}
    \bm{Z}_t = \text{GCN}\left(\left[\text{GCN}\left(\bm{X}_t, \bm{A}_t\right), \bm{h}_t\right], \bm{A}_t\right).
\label{eqn:latent}
\end{equation}
The fusion with $\bm{h}_t$ enables the latent relational representation aware of temporal contextual information. This fusion method is demonstrated to be effective to boost the performance of accident anticipation in our experiments.

\subsection{BNNs for Accident Anticipation}

\begin{figure}
    \centering
    \begin{minipage}[b]{0.48\textwidth}
      \centering
      \subfigure[Neural Networks]{
        \includegraphics[width=0.49\linewidth]{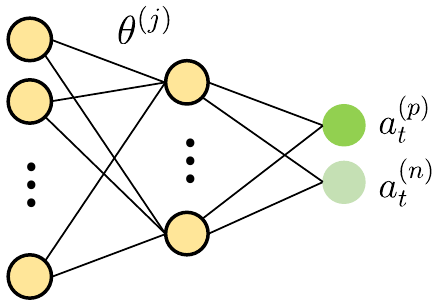}
        \label{fig:nn}
      }%
      \centering
      \subfigure[Bayesian Neural Networks]{
        \includegraphics[width=0.48\linewidth]{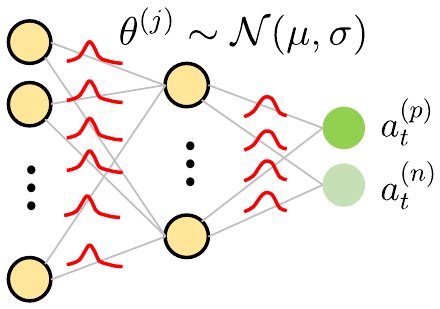}
        \label{fig:bnn}
      }%
    \end{minipage}
    \caption{Compared with NNs (Fig..~\ref{fig:nn}), network parameters of BNNs (Fig.~\ref{fig:bnn}) are sampled from Gaussian distributions so that both $\bm{a}_t$ and its uncertainty can be obtained.}
    \label{fig:comp_nns}
\end{figure}

To predict traffic accident score $\bm{a}_{t}$, a straightforward way is to utilize neural networks (NNs) as shown in Fig.~\ref{fig:nn}. However, the output of NNs is a point estimate which cannot address the intrinsic variability of the input relational features at each time step. Moreover, NNs could be overconfident in false model predictions when the model suffers from over-fitting problem.

To this end, we incorporate Bayesian neural networks (BNNs)~\cite{DenkerNIPS1990,NealBook2012} into our framework for accident score prediction. The architecture is shown in Fig.~\ref{fig:bnn}. The BNNs module consists of two BNN layers with latent representation $\bm{Z}_t$ given by Eq.~\ref{eqn:latent} as input to predict accident score $\bm{a}_{t}$. To best of our knowledge, we are the first to incorporate BNNs into video-based traffic accident anticipation such that predictive uncertainty can be achieved. The predictive uncertainty could be utilized to not only guide the relational features learning (see Section~\ref{sec:rankloss}), but also provide tools to interpret the model performance.

As we formulate the accident anticipation part as BNNs, the network parameters of BNNs such as weights and biases are all random variables, denoted as $\bm{\theta}$. Each entry of $\bm{\theta}$ is drawn from a Gaussian distribution determined by a mean and variance, i.e., $\bm{\theta}^{(j)} \sim \mathcal{N}(\mu, \sigma)$, in which $\bm{\alpha}^{(j)}=(\mu, \sigma)$ need to be learned with dataset $\mathcal{D}=(\bm{Z}_t, \bm{a}_t)$. Therefore, the likelihood of prediction can be expressed as $p(\bm{a}_t|\bm{Z}_t, \bm{\theta})=\mathcal{N}(f(\bm{Z}_t; \bm{\theta}), \beta)$, where $\beta$ is the predictive variance. However, according to Bayesian rule, to obtain the true posterior of model parameters, i.e., $p(\bm{\theta}|\mathcal{D})$, in addition to the likelihood and prior of $\bm{\theta}$, the marginal distribution $\int p(\bm{a}_t|\bm{Z}_t,\bm{\theta}) d\bm{\theta}$ is required, which is intractable since $\bm{a}_t=f(\bm{Z}_t, \bm{\theta})$ is modeled by a complex neural network. To estimate $p(\bm{\theta}|\mathcal{D})$, existing variational inference methods (VI)~\cite{GravesNIPS2011,BlundellJMLR2015,GalICLRW2016} could be used.

In this paper, we adopt the VI method Bayes-by-Backprop~\cite{BlundellJMLR2015} to approximate $p(\bm{\theta}|\mathcal{D})$ since it can be seamlessly incorporated in standard gradient-based optimization to learn from large-scale video dataset. According to~\cite{BlundellJMLR2015}, the variational approximation aims to minimize the following objective:
\begin{equation}
     \argmin_{\bm{\alpha}} \sum_{i=1}^{J}\log q\left(\bm{\theta}_i | \bm{\alpha}\right) - \log p\left(\bm{\theta}_i\right) - \log\left(p\left(\mathcal{D}|\bm{\theta}_i\right)\right),
\label{eqn:appro}
\end{equation}
where $J$ is the number of Monte Carlo samplings for $\bm{\theta}$. The first term  $q\left(\bm{\theta}_i | \bm{\alpha}\right)$ is the variational posterior distribution parameterized by $\bm{\alpha}$. The distribution parameters $\bm{\alpha}$ can be efficiently learned by using reparameterization trick and standard gradient descent methods~\cite{BlundellJMLR2015}. We denote this loss term as $\mathcal{L}_{VPOS}$. The second term $p(\bm{\theta}_i)$ is the prior distribution of $\bm{\theta}$. It is typically modeled with a spike-and-slab distribution, i.e., a mixture of two Gaussian density functions with zero means but different variances. We denote this loss term as $\mathcal{L}_{PRI}$.

The third term in Eq.~\ref{eqn:appro} is the negative log-likelihood of model predictions. Since minimizing this term is equivalent to minimizing the mean squared error (MSE), in this paper, we propose to use exponential binary cross entropy to achieve this objective:

\begin{equation}
    \mathcal{L}_{EXP} = \sum_{t=1}^{T}-e^{-\max \left(0, \frac{y-t}{f}\right)}\log a_t^{(p)} + \sum_{t=1}^{T}-\log \left(1 - a_t^{(n)}\right),
\label{eqn:exp}
\end{equation}
where $f$ is the constant frame rate for the given video, and $y$ is the beginning time of an accident provided by training set. The exponential weighted factor applies larger penalty to the time step that is closer to the beginning time of an accident.

\subsection{Uncertainty-guided Ranking Loss}
\label{sec:rankloss}

With the Bayesian formulation for accident anticipation, we can perform multiple forward passes at each time step such that an assembled prediction could be obtained by taking the average of these multiple outputs. Furthermore, as suggested by~\cite{KendallNIPS2017}, the predictive uncertainty (variance) can be decomposed as \textbf{aleatoric} uncertainty and \textbf{epistemic} uncertainty~\cite{KwonMIDL2018,ShridharArXiv2018}:
\begin{equation}
     \bm{U}_t = \underbrace{\frac{1}{M}\sum_{i=1}^{M} \left[\text{diag} \left(\hat{\bm{a}}_{i}\right) - \hat{\bm{a}}_{i} \hat{\bm{a}}_{i}^{T} \right]}_{\text{Aleatoric Uncertainty} (\bm{U}_{t}^{alt})} + \underbrace{\frac{1}{M} \sum_{i=1}^{M} \left(\hat{\bm{a}}_{i} - \bar{\bm{a}}\right) \left(\hat{\bm{a}}_{i} - \bar{\bm{a}}\right)^{T}}_{\text{Epistemic Uncertainty}(\bm{U}_{t}^{ept})},
\label{eqn:uncertain}
\end{equation}
where $\bar{\bm{a}} = \frac{1}{M} \sum_{i=1}^{M}\hat{\bm{a}}_{i}$ and $\hat{\bm{a}}_{i}=(\hat{a}_t^{(n)}, \hat{a}_t^{(p)})_i^T$. They are the predictions of the $i$-th forward pass at time step $t$ with total $M$ forward passes. The first term in Eq.~\ref{eqn:uncertain} is the aleatoric uncertainty, which measures the input variability (noise) of BNNs. In our model, the aleatoric uncertainty serves as an indicator to the quality of the learned relational features from GCNs and RNNs.

The second term in Eq.~\ref{eqn:uncertain} is epistemic uncertainty which is determined by the BNNs model itself. Inspired by Ma~\etal~\cite{MaCVPR2016}, ideally the epistemic uncertainties of sequential predictions should be monotonically decreasing, since as more frames the model observes, the more confident of the learned model (smaller epistemic uncertainty) will be. Therefore, we propose a novel ranking loss:
\begin{equation}
    \mathcal{L}_{RANK} = \max \left(0, \text{trace}\left(\bm{U}_t^{ept} - \bm{U}_{t-1}^{ept}\right)\right),
\label{eqn:rank}
\end{equation}
where $\bm{U}_{t-1}^{ept}$ and $\bm{U}_{t}^{ept}$ are epistemic uncertainties of successive frames $t-1$ and $t$ defined in Eq.~\ref{eqn:uncertain}. Note that $\bm{U}_t$ as well as the two terms in Eq.~\ref{eqn:uncertain} are matrices with size $2\times 2$, therefore in practice we propose to use matrix trace to quantify the uncertainties, which is similar to the method adopted in~\cite{ShridharArXiv2018}. Our proposed ranking loss aims to apply penalty to the predictions that do not follow the epistemic uncertainty ranking rule.

For aleatoric uncertainty $\bm{U}_{t}^{alt}$, it is not necessary to satisfy the monotonic ranking requirement since the noise ratio of accumulated data in video sequence is intrinsically not monotonic.

\begin{figure}
    \centering
    \includegraphics[width=\linewidth]{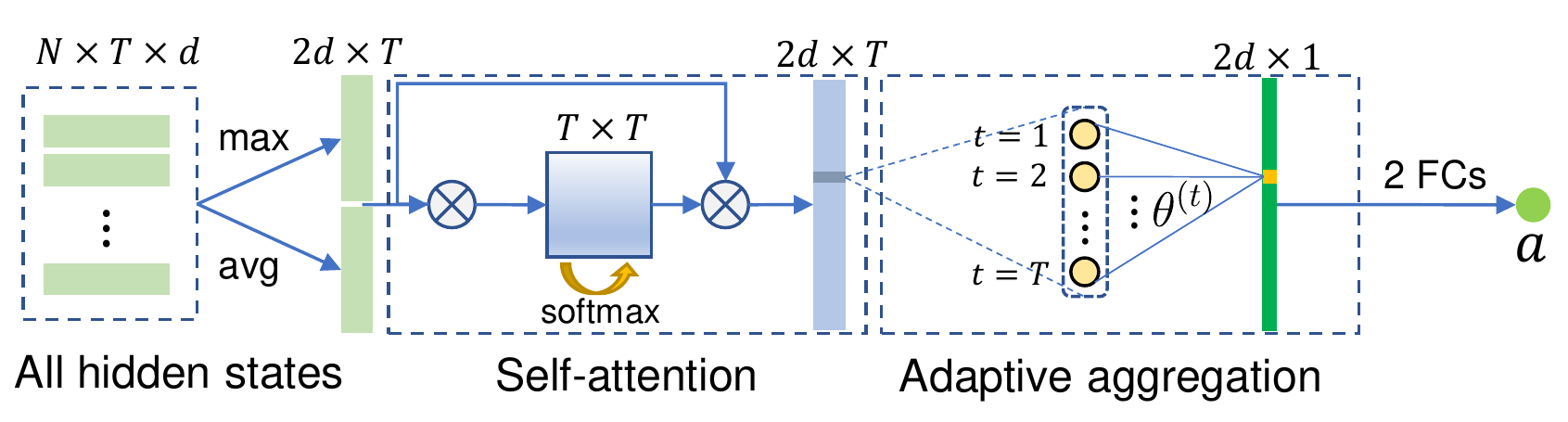}
    \caption{SAA Layer. First, all $N\times T$ hidden states are gathered and pooled by max-avg concatenation. Then, the simplified self-attention and adaptive aggregation are proposed to predict video-level accident score $\bm{a}$.}
    \label{fig:saa}
\end{figure}

\subsection{Temporal Self-Attention Aggregation}

Recurrent network can naturally build temporal relations of observations. However, the drawback of RNNs is that inaccurate hidden states in early temporal stages could be accumulated in iterative procedure and mislead the model to give false predictions in latter temporal stages. Besides, the hidden states in different time steps should be adaptive to anticipate the occurrence of a future accident.

To this end, motivated by recent self-attention design~\cite{VaswaniNIPS2017}, we propose a self-attention aggregation (SAA) layer in the training stage by adaptively aggregating hidden states of all time steps. Then, we use the aggregated representation to predict video-level accident score. The architecture of SAA layer is shown in Fig.~\ref{fig:saa}.

Specifically, we first aggregate hidden states of $N$ individual objects at each time step by applying the concatenation between mean- and max-pooling results. Then, the self-attention~\cite{VaswaniNIPS2017} is adapted to weigh the representation of all $T$ time steps. In this module, the embedding layers are not used. Lastly, instead of using simple average pooling, we introduce an FC layer with $T$ learnable parameters to adaptively aggregate the $T$ temporal hidden states. The aggregated video-level representation is used to predict the video-level accident score $\bm{a}$ by two FC layers. This network is trained with binary cross-entropy (BCE) loss:
\begin{equation}
    \mathcal{L}_{BCE} = -\log a^{(p)} - \log\left(1-a^{(n)}\right),
\end{equation}
where $\bm{a}=(a^{(n)},a^{(p)})^T$ normalized by softmax function. This auxiliary learning objective encourages the model to learn better hidden states even though SAA layer is not used in testing stage. 

Finally, the complete learning objective of our model is to minimize the following weighted loss:
\begin{equation}
    \mathcal{L} = \mathcal{L}_{EXP} + w_1 \cdot \left(\mathcal{L}_{VPOS} - \mathcal{L}_{PRI}\right) + w_2 \cdot \mathcal{L}_{RANK} + w_3 \cdot \mathcal{L}_{BCE}
\label{eqn:total_loss}
\end{equation}
where the $\mathcal{L}_{VPOS}$ and $\mathcal{L}_{PRI}$ are loss functions of variational posterior and prior. The constants $w_1$, $w_2$ and $w_3$ are set to 0.001, 10 and 10, respectively, to balance the magnitudes of these loss terms.
The second penalty term $\left(\mathcal{L}_{VPOS} - \mathcal{L}_{PRI}\right)$ is also termed as complexity loss and has similar effect to overcome over-fitting problem. The third penalty term $\mathcal{L}_{BCE}$ introduces video-level classification guidance while the fourth term $\mathcal{L}_{RANK}$ brings uncertainty ranking guidance to train our model.

\section{Experimental Results}

In this section, we evaluate our model on three real-world datasets, including our collected Car Crash Dataset (CCD) and two public datasets, i.e., Dashcam Accident Dataset (DAD)~\cite{ChanACCV2016} and AnAn Accident Detection (A3D) dataset~\cite{YaoIROS2019}. State-of-the art methods are compared and ablation studies are performed to validate our model.

\begin{figure*}
    \centering
    \includegraphics[width=0.98\textwidth]{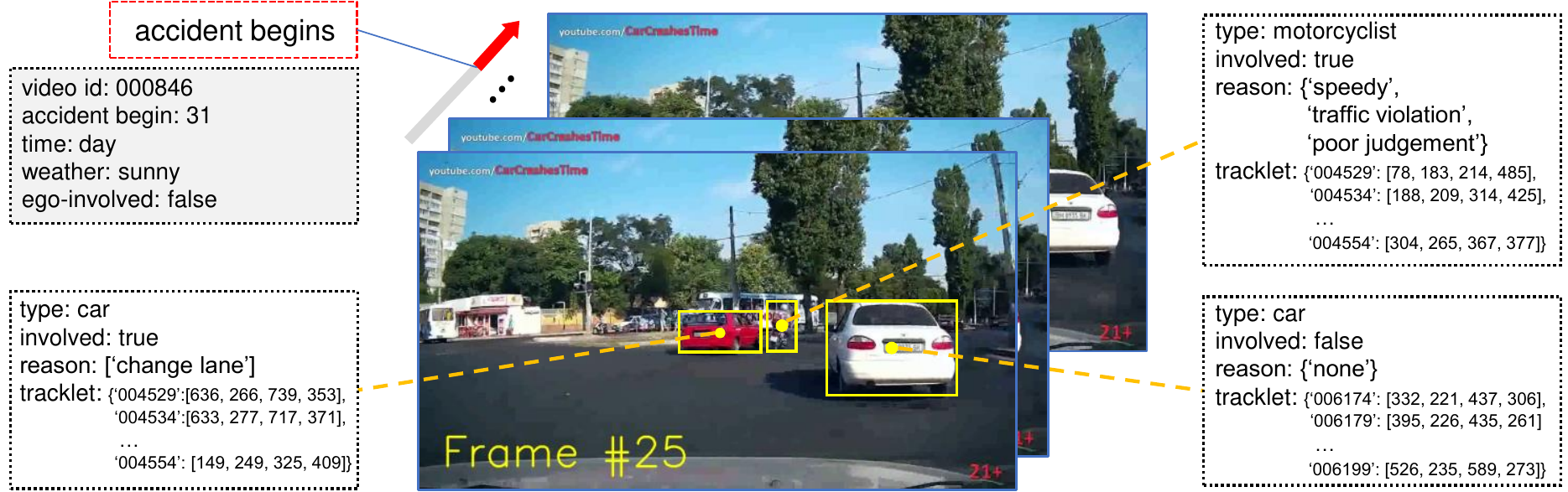}
    \caption{Annotation samples of our Car Crash Dataset (CCD). The gray box on top-left contains video-level annotations, while the other three white boxes provide instance-level annotations.
    }
    \label{fig:anno_vis}
\end{figure*}

\begin{table*}
\centering
\setlength{\tabcolsep}{0.6mm}
\caption{Comparison between CCD dataset and existing datasets. Information about DAD and A3D is obtained from their released sources. \emph{Temporal Annos.} means the temporal accident time annotations. \emph{Random ABT.} means accidents beginning times are randomly placed. \emph{Ego-involved} means the ego-vehicles are involved in accidents. \emph{Day/Night} indicates the data is collected in day or night. \emph{Weather} includes rainy, snowy, and sunny conditions. \emph{Participants Bbox} means the bounding boxes tracklets for accident participants. \emph{Accident Reasons} contains multiple possible reasons for each accident participant.}
\label{tab:data_comp}\small
\begin{tabular}{l|c|c|c|c|c|c|c|c|c|c}
\hline
Datasets & \# Videos & \# Positives & Total Hours & Temporal Annos. &Random ABT. &Ego-Involved &Day/Night &Weather &Participants Bbox &Accident Reasons \\
\hline
DAD~\cite{ChanACCV2016} &1,750 &620 &2.43 h &\checkmark & & & & & & \\
A3D~\cite{YaoIROS2019} &1,500 &1,500 &3.56 h &\checkmark &\checkmark &\checkmark & & & & \\
Ours (CCD) &4,500 &1,500 &6.25 h &\checkmark &\checkmark &\checkmark &\checkmark &\checkmark &\checkmark &\checkmark \\
\hline
\end{tabular}
\end{table*}

\subsection{Datasets}

\textbf{CCD dataset\footnote{CCD dataset is available at: \url{https://github.com/Cogito2012/CarCrashDataset}}.} In this paper, we collect a challenging Car Crash Dataset (CCD) for accident anticipation. We ask annotators to label YouTube accident videos with temporal annotations, diversified environmental attributes (day/night, snowy/rainy/good weather conditions), whether ego-vehicles involved, accident participants, and accident reason descriptions. For temporal annotations, the accident beginning time is labeled at the time when a car crash actually happens. To get trimmed videos with 5 seconds long, the accident beginning times are further randomly placed in last 2 seconds, generating 1,500 traffic accident video clips. We also collected 3,000 normal dashcam videos from BDD100K~\cite{YuCVPR2020} as negative samples. The dataset is divided into 3,600 training videos and 900 testing videos. Examples are shown in Fig.~\ref{fig:anno_vis} and comparison details with existing datasets are reported in Table~\ref{tab:data_comp}. Compared with DAD~\cite{ChanACCV2016} and A3D~\cite{YaoIROS2019}, our CCD is larger with diversified annotations.

\textbf{DAD dataset.} DAD~\cite{ChanACCV2016} contains dashcam videos collected in six cities in Taiwan. It provides 620 accident videos and 1,130 normal videos. Each video is trimmed and sampled into 100 frames with totally 5 seconds long. For accident videos, accidents are placed in the last 10 frames. The dataset has been divided into 1,284 training videos (455 positives and 829 negatives) and 466 testing videos (165 positives and 301 negatives).

\textbf{A3D dataset.} A3D~\cite{YaoIROS2019} is also a dashcam accident video dataset. It contains 1,500 positive traffic accident videos. 
In this paper, we only keep the 587 videos in which ego-vehicles are not involved in accidents. We sampled each A3D video with 20 fps to get 100 frames in total and placed the beginning time of each accident at the last 20 frames similar to DAD. The dataset is divided into 80\% training set and 20\% testing set.

\subsection{Evaluation Metrics}
\label{sec:eval}

\textbf{Average Precision.} This metric evaluates the \textbf{correctness} of identifying an accident from a video. Following the same definition as~\cite{ChanACCV2016}, at time step $t$, if $a_t^{(p)}$ is larger than a threshold, then the prediction at frame $t$ is positive to contain an accident, otherwise it is negative. For accident videos, all frames are labeled with ones (positive), otherwise the labels are zeros (negative). By this way, the precision, recall, as well as the derived Average Precision (AP) can be adopted to evaluate models.

\textbf{Time-to-Accident.} This metric evaluates the \textbf{earliness} of accident anticipation based on positive predictions. 
For a range of threshold values, multiple TTA results as well as corresponding recall rates can be obtained. Then, we use mTTA and TTA@0.8 to evaluate the earliness, where mTTA is the average of all TTA values and TTA@0.8 is the TTA value when recall rate is 80\%.
Note that if a large portion of predictions are false positives, very high TTA results can still be achieved while corresponding AP would be low. That means the model is overfitting on accident video and may give positive predictions for arbitrary input.
Therefore, except for fair comparison with existing methods, we mainly report TTA metrics when the highest AP is achieved, because it is meaningless to obtain high TTA if high AP cannot be guaranteed.

\textbf{Predictive Uncertainty.} Based on Eq.~\ref{eqn:uncertain}, we introduce to use the mean aleatoric uncertainty (mAU) and mean epistemic uncertainty (mEU) to evaluate the predictive uncertainties.

\subsection{Implementation Details}

We implement our model with PyTorch~\cite{pytorch}. For DAD dataset, we use the candidate objects and corresponding features provided by DSA\footnote{DSA: https://github.com/smallcorgi/Anticipating-Accidents} for fair comparison. For the experiments on A3D and CCD datasets, we use the public detection codebase MMDetection\footnote{MMDetection: https://github.com/open-mmlab/mmdetection} to train Cascade R-CNN~\cite{CaiCVPR2018} with ResNeXt-101~\cite{XieCVPR2017} backbone and FPN~\cite{LinCVPR2017} neck as our object detector on KITTI 2D detection dataset~\cite{Geiger2012CVPR}. The trained detector is used to detect candidate objects and then extract VGG-16 features of full-frame and all objects. 
As suggested by Bayes-by-backprop~\cite{BlundellJMLR2015}, we set the number of forward passes $M$ to 2 in training stage and 10 for testing stage. For the hyper-parameters of prior distribution, we set the mixture ratio $\pi$ to 0.5 and the variances of the two Gaussian distributions $\sigma_1$ to 1 and $\sigma_2$ to $\exp (-6)$. The dimensions of both hidden state of RNN and output of GCNs are set to 256. In the training stage, we set batch size to 10 and initial learning rate to 0.0005 with \emph{ReduceLROnPlateau} as learning rate scheduler. The model is trained by Adam optimizer for totally 70 training epochs.

\subsection{Performance Evaluation}

\begin{table}
\setlength{\tabcolsep}{1.5mm}
\centering
\caption{Evaluation results on DAD, A3D, and CCD datasets. Results of baselines on DAD are obtained from~\cite{ZengCVPR2017} and~\cite{SuzukiCVPR2018}. The notation ``--" means the metric is not applicable.}
\label{table:main}
\normalsize
\begin{tabular}{l|c|cccccc}
    \hline
    Datasets &Methods &mTTA(s) &AP(\%) &mAU &mEU \\
    \hline
    \multirow{4}{*}{DAD~\cite{ChanACCV2016}}
    &DSA~\cite{ChanACCV2016} & 1.34 & 48.1 & -- & -- \\  
    &L-RAI~\cite{ZengCVPR2017} & 3.01 & 51.4 & -- & -- \\
    &adaLEA~\cite{SuzukiCVPR2018} & 3.43 & 52.3 & -- & -- \\
    &Ours & \textbf{3.53} & \textbf{53.7} & \textbf{0.0294} & \textbf{0.0011} \\
    \hline
    \multirow{2}{*}{A3D~\cite{YaoIROS2019}}
    &DSA~\cite{ChanACCV2016} & 4.41 & 93.4 & -- & --\\
    &Ours & \textbf{4.92} & \textbf{94.4} & \textbf{0.0095} & \textbf{0.0023} \\
    \hline
    \multirow{2}{*}{CCD} &DSA~\cite{ChanACCV2016} & 4.52 & \textbf{99.6} & -- & -- \\
    &Ours & \textbf{4.74} & 99.5 & \textbf{0.0137} & \textbf{0.0001} \\
    \hline
\end{tabular}
\end{table}

\textbf{Compare with State-of-the-art Methods.} Existing methods~\cite{ChanACCV2016,ZengCVPR2017,SuzukiCVPR2018} are compared and results are reported in Table~\ref{table:main}. For fair comparison, we use the model at the last training epoch for evaluation on DAD datasets. Nevertheless, the trained model with best AP is kept for evaluation on other two datasets since high AP is important to suppress impact of false positives on TTA evaluation. Note that these two metrics currently are only applicable to our model, since we are the first to introduce uncertainty formulation for accident anticipation.

From Table~\ref{table:main}, our model on DAD dataset achieves the best mTTA which means the model anticipates on average 3.53 seconds earlier before an accident happens, while keeping competitive AP performance at 53.7\% compared with L-RAI and adaLEA. Note that the video lengths of the three datasets are all 5 seconds, our high performance on A3D and CCD demonstrate that our model is easier to be trained on different datasets. This can be explained by the mAU results due to their consistence with TTA evaluation results in Table~\ref{table:main}. The low mAU values on A3D and DAD datasets reveal that our model has learned relational representations with high quality on these datasets.

We further report TTA results with different recall rates from 10\% to 90\% in Table~\ref{tab:ttar}. It shows that our model outperform DSA in most of recall rate requirements. For recall rates larger than 80\%, our method performs poorly compared with DSA. However, high recall rate may also lead to too much false alarm so that AP cannot be guaranteed to be high. This finding also supports our motivation to use the trained model with best AP for evaluation.

\begin{figure*}[t]
    \centering
    \includegraphics[width=\textwidth]{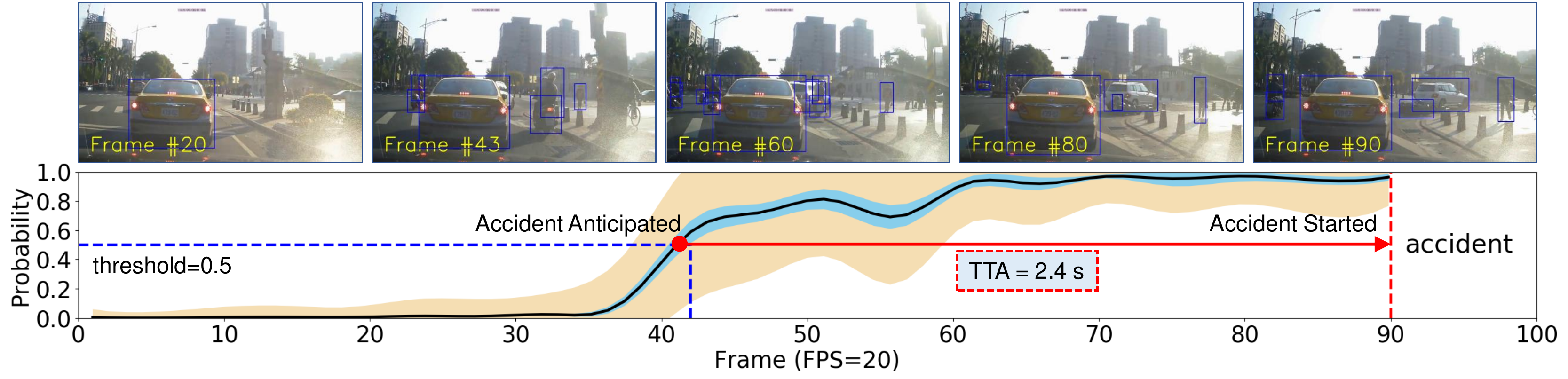}
    \includegraphics[width=\textwidth]{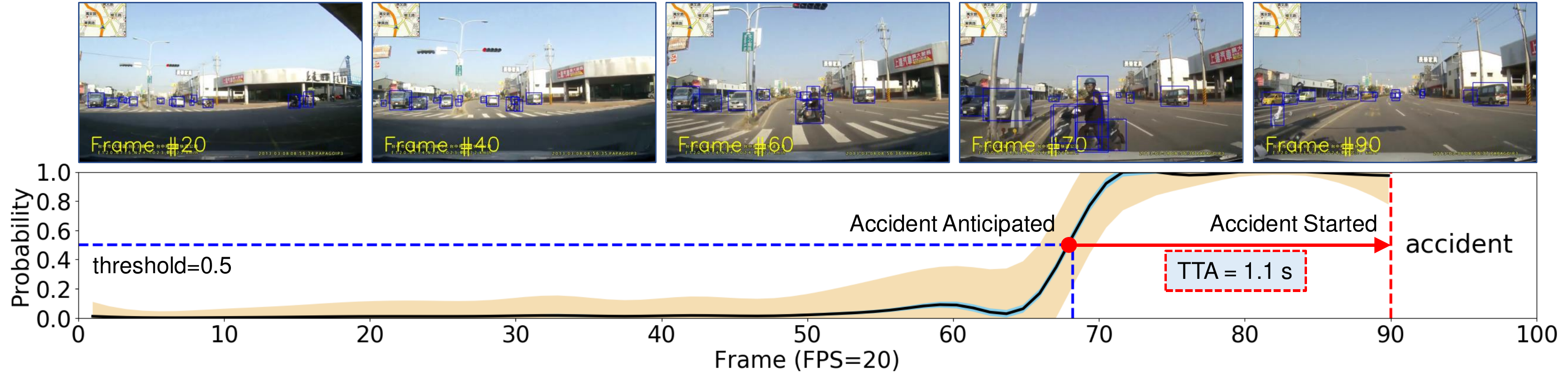}
    \includegraphics[width=\textwidth]{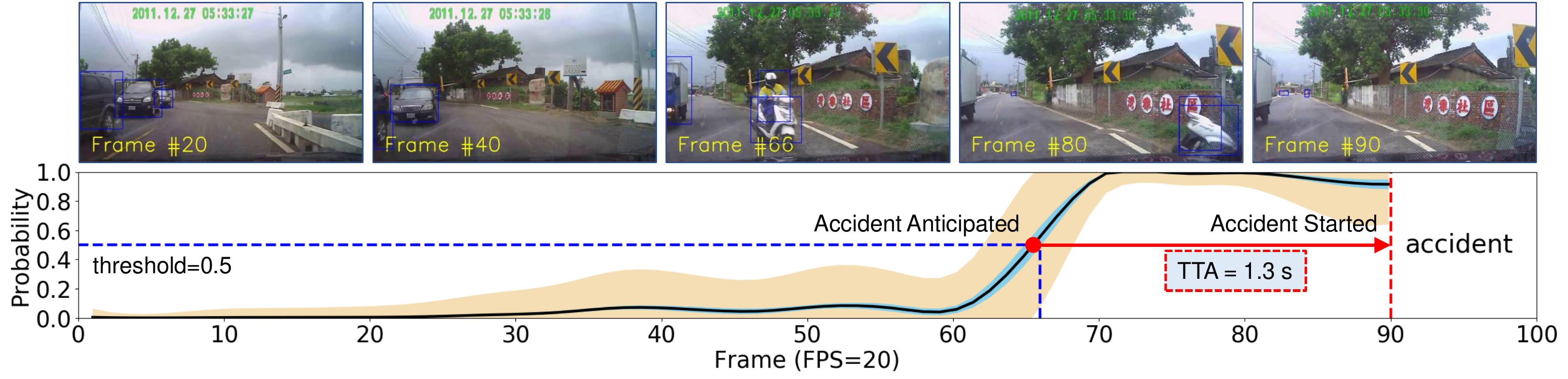}
    \caption{Examples of our predictions on DAD datasets. The red curves indicate smoothed accident scores as observed frames increase. The ground truth (beginning time of accident) are labeled at 90-th frame. We plot one time of squared epistemic (blue region) and aleatoric uncertainties (wheat color region). The horizontal line indicates probability threshold 0.5.}
    \label{fig:vis}
\end{figure*}

\textbf{Visualization} We visualized accident anticipation results with samples in DAD dataset (see Fig.~\ref{fig:vis}). The uncertainty regions indicate that in both early and late stages, the model is quite confident on prediction (low uncertainties), while in the middle stage when accident scores start are increasing, the model is uncertain to give predictions. Note that the predicted epistemic uncertainty (blue region) is not necessary to be monotonically decreasing since we only use Eq.~\ref{eqn:rank} as training regularizer rather than strict guarantee on predictions. The results are with good interpretability, in that driving system is typically quite sure about the accident risk level when the self-driving car is far from or almost being involved in an accident, while it is uncertain about it when accumulated accident cues are insufficient to make decision.

\begin{table}[t]
\centering
\caption{TTA with different recall rates on DAD dataset.}
\setlength{\tabcolsep}{1.0mm}
\label{tab:ttar}
\normalsize
\begin{tabular}{l|ccccccccc}
\hline
Recall &0.1 &0.2 &0.3 &0.4 &0.5 &0.6 &0.7 &0.8 &0.9 \\
\hline
DSA~\cite{ChanACCV2016} &0.28 &0.50 &0.73 &0.87 &0.92 &1.02 &1.24 &1.35 &\textbf{2.28} \\
Ours &\textbf{0.59} &\textbf{0.75} &\textbf{0.84} &\textbf{0.96} &\textbf{1.07} &\textbf{1.16} &\textbf{1.33} &\textbf{1.56} &1.99 \\
\hline
\end{tabular}
\end{table}

\subsection{Ablation Study}

In this section, to validate the effectiveness of the several main components, the following components are replaced or removed, and compared with our model based on best AP setting. (1) \textbf{BNNs}: The BNNs are replaced with vanilla FC layers. Note that in this case, $\mathcal{L}_{VPOS}-\mathcal{L}_{PRI}$ and our proposed ranking loss $\mathcal{L}_{RANK}$ in Eq.~\ref{eqn:total_loss}, as well as mAU are not applicable. (2) \textbf{SAA}: The SAA layer is removed so that $\mathcal{L}_{BCE}$ in Eq.~\ref{eqn:total_loss} is not used. (3) \textbf{GCN}: We replace GCNs with vanilla FC layers in Eq.~\ref{eqn:rnn} and Eq.~\ref{eqn:latent}. (4) \textbf{Fusion}: For this variant, the fusion in Eq.~\ref{eqn:rnn} and Eq.~\ref{eqn:latent} are removed such that only $\bm{Z}_t$ and $\text{GCN}(\bm{X}_t,\bm{A}_t)$ are used, respectively. (5) \textbf{RankLoss}: The epistemic uncertainty-based ranking loss is removed so that $\mathcal{L}_{RANK}$ in Eq.~\ref{eqn:total_loss} is not applicable. Results are shown in Table~\ref{tab:ablation}. 

\begin{table}[t]
\centering
\setlength{\extrarowheight}{0.5mm}
\setlength{\tabcolsep}{0.8mm}
\caption{Ablation studies results on DAD dataset. 
}
\label{tab:ablation}
\normalsize
\begin{tabular}{c|c|c|c|c|c|c|c}
\hline
    Variants &BNNs &SAA &GCN &Fusion &RankLoss &AP(\%) &mAU \\
\hline
    (1) &\checkmark  &\checkmark &\checkmark &\checkmark &\checkmark & \textbf{72.22} & \textbf{0.0731} \\
    (2) & &\checkmark &\checkmark &\checkmark & & 70.38 & -- \\
    (3) &\checkmark  & &\checkmark &\checkmark &\checkmark & 67.34 & 0.1150\\
    (4) &\checkmark  &\checkmark & &\checkmark &\checkmark & 67.10 & 0.1250\\
    (5) &\checkmark  &\checkmark &\checkmark & &\checkmark & 65.50 & 0.1172\\
    (6) &\checkmark  &\checkmark &\checkmark &\checkmark & &  64.60 & 0.0950\\
\hline
\end{tabular}
\end{table}

We can clearly see that the uncertainty-based ranking loss contributes most to our model by comparing variant (2)(6) with (1), with about 7.6\% performance gain. Though the BNNs module leads to small performance gain, we attribute the benefit of BNNs to its derived uncertainty ranking loss as well as the interpretable results. Furthermore, the lowest mAU and highest AP for variant (1) demonstrate that the learned relational features are of highest quality (smallest uncertainty) compared with other variants.

\begin{table}
\centering
\setlength{\tabcolsep}{3.0mm}
\caption{Model size comparison. Our model variants (2), (4), and (5) are included for comparison. Unit M means a million.}
\label{tab:size}
\normalsize
\begin{tabular}{l|c|c|c|c|c}
\hline
Methods &DSA &Ours & v(2) & v(4) & v(5) \\
\hline
\# Params. (M) &4.40 &1.97 &1.66 &1.97 &1.90 \\
\hline
\end{tabular}
\end{table}

The results of variants (3) validate the effectiveness of our self-attention aggregation (SAA) layer, while the results of variant (4) validate the superiority GCN over naive FC layers. The results of variant (5) show that the feature fusion between GCN outputs and hidden states, and the fusion between relational features and agent-specific features are important to accident anticipation, leading to approximately $7\%$ performance gain.

\textbf{Model Size Comparison.} The number of network parameters are counted and reported in Table~\ref{tab:size}. It shows that the proposed model is much light-weighted than DSA, and only slightly increases the model size when compared with other variants of our model.

\section{Conclusion}

In this paper, we propose an uncertainty-based traffic accident anticipation with spatio-temporal relational learning. Our model can handle the challenges of relational feature learning and anticipation uncertainty from video data. Moreover, the introduced Bayesian formulation not only significantly boosts anticipation performance by using the uncertainty-based ranking loss, but also provides interpretation on predictive uncertainty. In addition, we release a Car Crash Dataset (CCD) for accident anticipation which contains rich environmental attributes and accident reason annotations.

\begin{acks}
 We thank NVIDIA for GPU donation and Haiting Hao for organizing dataset collection. This research is supported by an ONR Award N00014-18-1-2875. The views and conclusions contained in this paper are those of the authors and should not be interpreted as representing any funding agency.
\end{acks}

\clearpage
\balance
\bibliographystyle{ACM-Reference-Format}
\bibliography{camera_ready}



\end{document}